\documentclass{article} 
\usepackage{iclr2026_conference,times}

\usepackage{amsmath,amsfonts,bm}

\def\eqref#1{equation~\ref{#1}}

\def\1{\bm{1}}

\def\vh{{\bm{h}}}

\def\vv{{\bm{v}}}

\DeclareMathAlphabet{\mathsfit}{\encodingdefault}{\sfdefault}{m}{sl}
\SetMathAlphabet{\mathsfit}{bold}{\encodingdefault}{\sfdefault}{bx}{n}

\usepackage{hyperref}
\usepackage{url}

\usepackage[pdftex]{graphicx}
\usepackage{subcaption}
\usepackage{booktabs}
\usepackage{makecell}
\usepackage{pifont}
\usepackage{kotex}
\usepackage{multirow}
\usepackage{float}
\usepackage{xspace}
\usepackage{wrapfig}
\usepackage{listings}
\usepackage{xcolor}
\usepackage{amssymb}     
\usepackage{sidecap}
\usepackage{amsmath}
\usepackage[most]{tcolorbox}
\definecolor{lightblue}{RGB}{220, 230, 255}
\definecolor{lightred}{RGB}{255, 220, 220}

\newcommand{\increase}[1]{
  \tcbox[
    on line, colback=lightblue, colframe=lightblue, boxrule=0pt,
    arc=1mm,        
    boxsep=0.7pt, 
    top=0.7pt,      
    bottom=0.7pt,
    left=1.5pt,   
    right=1.5pt,
    nobeforeafter
  ]{\textcolor{blue}{\scriptsize$\mathbf{\uparrow#1}$}}
}

\newcommand{\decrease}[1]{
  \tcbox[
    on line, colback=lightred, colframe=lightred, boxrule=0pt,
    arc=1mm,       
    boxsep=0.7pt,   
    top=0.7pt,    
    bottom=0.7pt,
    left=1.5pt,   
    right=1.5pt,
    nobeforeafter
  ]{\textcolor{red}{\scriptsize$\mathbf{\downarrow#1}$}}
}

\usepackage{marvosym}
\newcommand*\colourcheck[1]{%
  \expandafter\newcommand\csname #1check\endcsname{\textcolor{#1}{\Circpipe}}
}
\definecolor{myred}{RGB}{220, 50, 50}

\colourcheck{green}

\lstdefinestyle{mypromptstyle}{
    basicstyle=\ttfamily\scriptsize,
    breaklines=true,
    captionpos=b,
    frame=single,
    rulecolor=\color{gray},
    keepspaces=true,
}

\title{Rethinking Cross-lingual Alignment:\\ Balancing Transfer and Cultural Erasure in Multilingual LLMs}

\stepcounter{footnote}\stepcounter{footnote}
\author{HyoJung Han\thanks{Work done at Google.}\\
University of Maryland\\
\texttt{\small{hjhan@cs.umd.edu}}
\And
Sweta Agrawal\\
Google\\
\texttt{\small{swetaagrawal@google.com}}
\And
Eleftheria Briakou\\
Google\\
\texttt{\small{ebriakou@google.com}}
}

\iclrfinalcopy 

\newif\ifcomment\commenttrue
\ifcomment
    \newcommand{\customcmt}[3]{\textcolor{#1}{[#2: #3]}}
\else
    \newcommand{\customcmt}[3]{}
\fi

\newcommand{\pca}{\textsc{pca}\xspace}
\newcommand{\cla}{\textsc{cla}\xspace}
\newcommand{\llm}{\textsc{llm}\xspace}
\newcommand{\llms}{\textsc{llm}s\xspace}
\newcommand{\gmmlu}{\textsc{gmmlu}\xspace}
\newcommand{\blend}{\textsc{blend}\xspace}
\newcommand{\base}{\textsc{unaligned}\xspace}
\newcommand{\mist}{\textsc{mist}\xspace}
\newcommand{\clo}{\textsc{clo}\xspace}
\newcommand{\midalign}{\textsc{midalign}\xspace}
\newcommand{\ensteer}{\textsc{en}-steering\xspace}
\newcommand{\locsteer}{\textsc{loc}-steering\xspace}
\newcommand{\sursteer}{\textsc{sur}-steering\xspace}

\newcommand{\tradeoff}{transfer-localization\xspace}

\begin{document}

\maketitle

\begin{abstract}
Cross-lingual alignment (\cla) aims to align multilingual representations, enabling Large Language Models (\llms) to seamlessly transfer knowledge across languages. While intuitive, we hypothesize, this pursuit of representational convergence can inadvertently cause ``cultural erasure''---the functional loss of providing culturally-situated responses that should diverge based on the query language. In this work, we systematically analyze this trade-off by introducing a holistic evaluation framework, the \tradeoff plane, which quantifies both desirable knowledge transfer and undesirable cultural erasure.
Using this framework, we re-evaluate recent \cla approaches and find that they consistently improve factual transfer at the direct cost of cultural localization across all six languages studied. Our investigation into the internal representations of these models reveals a key insight: universal factual transfer and culturally-specific knowledge are optimally steerable at different model layers.
Based on this finding, we propose Surgical Steering, a novel inference-time method that disentangles these two objectives. 
By applying targeted activation steering to distinct layers, our approach achieves a better balance between the two competing dimensions, effectively overcoming the limitations of current alignment techniques.
\end{abstract}

\section{Introduction}
\label{sec:intro}

\begin{figure}[!t]
    \centering
    \includegraphics[width=0.85\linewidth]{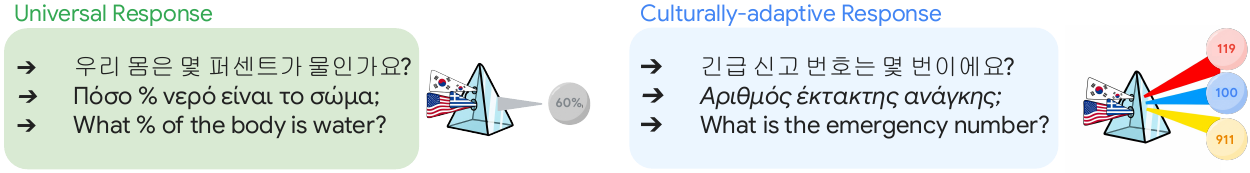}
    \caption{Examples of intended convergence and desired divergence in outputs of multilingual \textsc{llm}s. Universal questions (left) should result in a single, converged answer (knowledge transfer) regardless of the query languages, while culturally-specific questions (right) should result in divergent, localized answers (cultural localization) reflecting cultural context inferred from the input language.}
    \label{fig:main_fig}
\end{figure}

Multilingual Large Language Models (\textsc{llm}s) are expected to perform knowledge transfer uniformly across all languages \citep{li-etal-2024-land, lu2025learnunlearnmisinfo}, transcending the inherent asymmetries in their training data~\citep{vahid2023detxlinginfogapwiki}. For example, a model that acquires knowledge in English for the question, \textit{``What \% of the body is water?''} should ensure that this knowledge is equally retrievable regardless of the query language. However, empirical studies have reported significant performance gaps across languages in multilingual tasks \citep{qi-etal-2023-clconsistency, jiang-etal-2020-xfactr, kassner-etal-2021-multilinglama}. To overcome these inconsistencies, multilingual \textsc{llm}s rely on cross-lingual alignment, aiming at bringing different language representations closer together. 
Within this framing, inconsistencies across languages are typically regarded as undesirable \citep{jiang-etal-2020-xfactr,ohmer2023sepformnmeaning}. 
However, this pursuit of uniformity creates a critical tension: what happens to knowledge that should be local? Consider the question (Figure \ref{fig:main_fig}): \textit{``What is the emergency number?''} Does representational alignment cause the model to default to \textit{``911''} regardless of the query language?  

While prior work on cross-lingual alignment (\textsc{cla}) has predominantly focused on its benefits for knowledge transfer, potential side effects remain underexplored. We address this gap by investigating a critical trade-off: the desirable transfer gained through alignment versus the undesirable loss of the model's ability to provide culturally localized responses.
In doing so, we ask the following questions:

\paragraph{How can we evaluate both the gains and losses of alignment?} 
We propose a holistic evaluation framework built on a two-dimensional \tradeoff plane (Section~\ref{sec:framework}). The first axis measures desirable transfer, where a model should provide consistent responses across languages. The second axis measures cultural localization, the model's ability to tailor its responses to the cultural context inferred from the input language. Within this plane, we identify an undesirable quadrant where high transfer is achieved at the cost of cultural erasure—a regression in the model's ability to adapt.

\paragraph{What hidden cultural costs accompany current cross-lingual alignment methods?} 
We re-evaluate a series of popular \textsc{cla} methods on the \tradeoff plane (Section~\ref{sec:alignment_cost}) and show that while these methods improved knowledge transfer, they consistently degrade the model’s ability to answer culturally specific questions, exposing a significant hidden cost.

\paragraph{How can we design culturally-aware alignment techniques to better balance the trade-off?}
By analyzing the model's internal representations, we identify a key distinction in how knowledge is encoded: while cross-lingual transfer is better realized within a model's middle layers, \textit{cultural localization is predominantly encoded in the deeper layers}. Leveraging this insight, we introduce a simple, layer-specific intervention to steer the model towards both universal and local subspaces (Section~\ref{sec:mitigating}). We show this method improves both transfer and localization across all \cla techniques, pushing performance into the desirable quadrant. Nevertheless, the trade-off is not fully eliminated, indicating that a residual loss of cultural nuance is inherent to the alignment process.

In summary, our work reframes the study of cross-lingual alignment by centering the critical trade-off between knowledge transfer and cultural localization, paving the way for the development of culturally-aware alignment in truly multilingual \textsc{llm}s.

\section{Related Work}

\paragraph{The Root of Multilingual Gaps: Data and Representational Asymmetry}

Performance gaps in multilingual models are often attributed to severe imbalances in their training data \citep{vahid2023detxlinginfogapwiki}. This asymmetry leads to a model where knowledge is primarily encoded in the representations of high-resource languages (like English), which dominate pre-training corpora \citep{wenzek-etal-2020-ccnet, pfeiffer2022curseofmling}.
Internally, this manifests as \textsc{llm}s processing multilingual inputs by mapping them to a shared, language-agnostic semantic space—one that is often heavily biased towards English—before translating them back to the target language for the final output \citep{zhao2024how, wendler-etal-2024-llamas, dumas2025separatingtongue}. Consequently, the degree of alignment between English and non-English representations has become a reliable proxy for multilingual capability \citep{kargaran-etal-2025-mexa, ravisankar2025mapeng}, while performance degradation of non-English is often linked to failures in this internal convergence or translation process \citep{wang-etal-2025-lost-multilinguality}.

\paragraph{Closing Multilingual Gaps: Cross-lingual Alignment}

\cla approaches have introduced throughout the \textsc{llm} development cicle. 
\textit{During pre-training}, alignment is implicitly induced as a byproduct of training on parallel data, which act as cross-lingual representation anchors\citet{blum2025rosettastoneunification}.
\textit{At post-training}, alignment is enhanced through multilingual instruction-tuning~\citep{ouyang2022sft,lai-etal-2023-okapi, zhang-etal-2024-self-distill}, or by introducing objectives that explicitly encourage semantic alignment or language-agnostic retrieval \citep{lee-etal-2025-cross, liu-niehues-2025-middle}.
\textit{At inference time}, proposed \cla include steering representations towards English \citep{lim2025langlatenthinders, lu2025pathstaken},
merging task and language-specific adapters
\citet{zhao-etal-2025-adamergex}, swapping layers between specialized models~\citet{bandarkar2025layerswap}, or simply  translating queries into English externally \citep{banea-etal-2008-translatetest, etxaniz-etal-2024-thinkbetteren}  or using cross-lingual thought prompting \citep{huang-etal-2023-clthoughtprompt}.

\paragraph{Culturally-Situated LLMs: Desired Representation Localization}

Cultural localization has become a central challenge for \textsc{llm}s, with recent research establishing they exhibit a strong Western-centric bias~\citep{bayramli-etal-2025-diffusion, zhou-etal-2025-mapo}. %
In response, major research efforts focus on creating benchmarks to diagnose %
these biases, by curating multilingual datasets \citep{clark-etal-2020-tydi, Salazar2025KaleidoscopeIE, hasan-etal-2025-nativqa}; investigating social constructs through datasets on stereotypes \citep{bhutani-etal-2024-seegull}; social norms \citep{forbes-etal-2020-social, rao-etal-2025-normad}; or divergent cross-lingual perspectives on the same topics~\citep{shwartz-2022-good, li-etal-2024-bordirlines}. Finally, recent work explores inference-time, culturally-aware approaches based on static methods~\citep{arora-etal-2023-probingcrossculdiff, lertvittayakumjorn-etal-2025-towards, li-etal-2024-land} or adapted prompting with agents \citep{ki-etal-2025-multiple}.

To date, research on \textbf{cross-lingual transfer and culturally-situated models has largely proceeded in isolation}, with the former focusing on enforcing cross-lingual representation alignment and the latter on localization rooted in cultural context.
We unify these two strands with a framework (Section \ref{sec:framework}) designed to uncover the hidden costs of alignment (Section \ref{sec:alignment_cost}) and to develop interventions that balance shared knowledge with cultural specificity  (Section \ref{sec:mitigating}).

\section{Measuring the \tradeoff Trade-off}\label{sec:framework}

In this work, we propose a framework that measures both the benefits of cross-lingual representation alignment and the costs of losing cultural localization nuance during \textsc{cla}. 
To formalize this, we introduce a typology of two distinct knowledge categories: %
\textit{universal knowledge} refers to language-invariant knowledge, where a model's response should remain (semantically) consistent across languages. Conversely, 
\textit{culturally-adaptive knowledge} is based on universal concepts but instantiated differently through local norms, cultural contexts, or regulations. In such scenarios, a model should preserve language-specific nuances, making output localization the intended behavior.

Building on this typology, we define two key metrics to evaluate \cla along a \textbf{\tradeoff} plane. As seen below, both metrics are defined as relative changes in performance compared to an unaligned baseline model, allowing us to precisely measure the impact of each alignment technique:

\begin{itemize}
    \item[\raisebox{-0.5ex}{\includegraphics[width=2.1em]{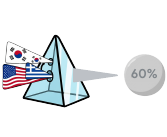}}] \textbf{Transfer}: We define transfer as the performance difference on \textit{universal} knowledge tasks after applying an alignment method. It quantifies the \textit{desirable} outcome of alignment: bridging the knowledge gap across different languages. A positive transfer score indicates that the model has successfully generalized knowledge from one language to another.
    \item[\raisebox{-0.5ex}{\includegraphics[width=1.8em]{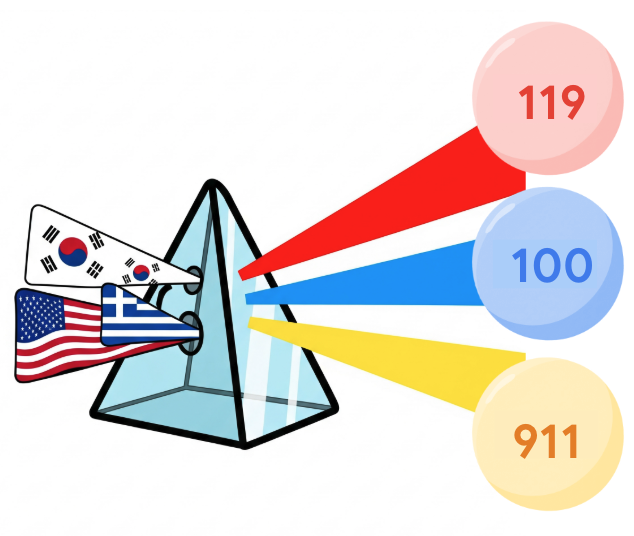}}] \textbf{Localization}: We define cultural localization
    as the performance difference on \textit{culturally-adaptive} tasks. A negative score indicates cultural erasure, representing a functional loss in the model's ability to handle culturally specific questions.
\end{itemize}

To \textbf{operationalize} this framework, we employ benchmarks tailored to each knowledge type. We quantify knowledge transfer using Global \textsc{mmlu} \citep[\gmmlu]{singh-etal-2025-global}, which contains universal multiple-choice questions across various academic and professional subjects. We measure cultural localization using a multilingual version of the \blend benchmark \citep{myung2024blend}, which is designed to evaluate knowledge of culturally and regionally specific concepts.\footnote{
To tailor \blend to our needs, we automatically generate a decontextualized version by removing explicit localization context (e.g., ``in Greece'') from its questions (details in Appenfix~\ref{sec:appendix_detail_benchmark}). This lets us test the model's ability to provide culturally-situated responses by inferring the right context from the language itself.} By plotting the change in \gmmlu accuracy (Transfer) against the change in \blend accuracy (Localization), we can map each \cla method to a point on the \tradeoff frontier, visualizing its trade-off.

\section{Uncovering the Hidden Cost of Alignment}\label{sec:alignment_cost}

We revisit a series of recent \cla approaches and evaluate them under our proposed \tradeoff framework. First, we discuss preliminaries of the studied \cla techniques (\S\ref{sec:prelim}), then describe our experimental setup (\S\ref{sec:setup}), and conclude with deep dives into the results (\S\ref{sec:mapping}-\S\ref{sec:cla_dynamics}).

\subsection{Cross-Lingual Alignment  Preliminaries}
\label{sec:prelim}

We focus on four recent \cla methods that have been proposed to foster representation alignment across languages. 
Some methods achieve this implicitly by training on parallel data, while others explicitly enforce alignment by directly manipulating or optimizing model representations, often guiding them toward English latent subspaces.  We consider a spectrum of approaches, covering two main paradigms: post-training and inference-time steering approaches which we detail below.

\paragraph{Multilingual Instruction Tuning (\mist)}employs 
a standard negative log likelihood (\textsc{nll}) loss, which involves training on multilingual datasets of query-response pairs. The multilingual datasets are usually derived by extending English datasets through translation and training on such data is shown to enhance a model's generalization capabilities across various languages. In this case, representation alignment is implicitly enforced as a byproduct of training on parallel instruction tuning datasets \citep{lai-etal-2023-okapi, blum2025rosettastoneunification}.

\paragraph{Middle-Layer Representation Alignment (\midalign)}introduces a more explicit alignment mechanism, alternating between a supervised fine-tuning (\textsc{sft}) loss and a dedicated cross-lingual alignment loss \citep{liu-niehues-2025-middle}. 
Concretely, activations from the middle layer~$\ell$ of the network are extracted for parallel texts ($\vh_{\text{\textsc{src}}}^{\ell}$, $\vh_{\text{\textsc{tgt}}}^{\ell}$) and mean-pooled over sequence.
The alignment loss ($\mathcal{L}_{\text{\textsc{midalign}}}$, Eq.~\ref{eq:midalign}) is then formulated to maximize the similarity between translations, while minimizing the cosine similarity between non-translations within the same batch~$\mathcal{B}$, which directly shapes the latent space to be more language agnostic. The loss is given as:

\begin{equation}
\mathcal{L}_{\text{\textsc{midalign}}} 
= - \log \frac{\exp\!\left(\cos(\vh_{\text{\textsc{src}}}^{\ell}, \vh_{\text{\textsc{tgt}}}^{\ell})\right)}
{\sum_{{\text{b}} \in \mathcal{B}} \exp\!\left(\cos(\vh_{\text{\textsc{src}}}^{\ell}, \vh_{\text{b}}^{\ell})\right)}.
\label{eq:midalign}
\end{equation}

\paragraph{Cross-lingual Optimization (\clo)} 
 aims at transferring an \textsc{llm}'s English capabilities to a target language by using a Cross-Lingual (\textsc{cl}) loss \citep{lee-etal-2025-cross}---an adaptation of the Direct Preference Optimization objective \citep{rafailov2023direct}. Concretely, for a non-English query~$x_{\text{\textsc{xx}}}$, English responses~$y_{\text{\textsc{en}}}$ are suppressed, while in-language responses~$y_{\text{\textsc{xx}}}$ are preferred, and vice versa; enabling the model to leverage its existing English knowledge for generating outputs in a target language. Formally, the loss is $\mathcal{L}_{\text{CLO}} = \lambda \, \mathcal{L}_{\text{SFT}} + (1 - \lambda) \, \mathcal{L}_{\text{CL}}$ where $\mathcal{L}_{\text{SFT}}$ is applied on non-English query-response pair ($x_{\text{\textsc{xx}}}$, $y_{\text{\textsc{xx}}}$), and $\mathcal{L}_{\text{CL}}$ is given as follows:
\begin{equation}
\mathcal{L}_{\text{CL}} 
= - \mathbb{E}_{(x_{\text{\textsc{en}}}, y_{\text{\textsc{en}}}, y_{\text{\textsc{xx}}}) \sim \mathcal{D}}
\!\left[ \log \sigma(z_{\text{\textsc{en}}}) \right]
- \mathbb{E}_{(x_{\text{\textsc{xx}}}, y_{\text{\textsc{xx}}}, y_{\text{\textsc{en}}}) \sim \mathcal{D}}
\!\left[ \log \sigma(z_{\text{\textsc{xx}}}) \right], \quad \text{where} \nonumber
\end{equation}
\begin{equation}
z_{\text{\textsc{en}}} 
= \beta \left( 
\log \tfrac{\pi_{\theta}(y_{\text{\textsc{en}}} \mid x_{\text{\textsc{en}}})}{\pi_{\text{ref}}(y_{\text{\textsc{en}}} \mid x_{\text{\textsc{en}}})}
- \log \tfrac{\pi_{\theta}(y_{\text{\textsc{xx}}} \mid x_{\text{\textsc{en}}})}{\pi_{\text{ref}}(y_{\text{\textsc{xx}}} \mid x_{\text{\textsc{en}}})}
\right), \quad
z_{\text{\textsc{xx}}} 
= \beta \left( 
\log \tfrac{\pi_{\theta}(y_{\text{\textsc{xx}}} \mid x_{\text{\textsc{xx}}})}{\pi_{\text{ref}}(y_{\text{\textsc{xx}}} \mid x_{\text{\textsc{xx}}})}
- \log \tfrac{\pi_{\theta}(y_{\text{\textsc{en}}} \mid x_{\text{\textsc{xx}}})}{\pi_{\text{ref}}(y_{\text{\textsc{en}}} \mid x_{\text{\textsc{xx}}})}
\right).
\end{equation}

\paragraph{English Steering (\ensteer)} is an inference-time intervention based on contrastive activation addition~\citep{rimsky-etal-2024-steering}, where ``steering vectors'' are computed to shift the model's distribution towards a desired behavior.
In the context of \cla, \cite{lim2025langlatenthinders} propose to shift a model's latent space towards English
motivated by prior work's observation that the shared latent space in multilingual \textsc{llm}s is closer to English \citep{wendler-etal-2024-llamas}.
Following this, we sample  contrastive pairs~$\mathcal{S}$ consisting of English and non-English parallel queries ($x_{\text{\textsc{en}}}$, $x_{\text{\textsc{xx}}}$). We then compute the average differences between the activations~$\vh^{\ell}(x)$ at layer~$\ell$ over all pairs, resulting in an English steering vector, $\vv_{\text{\textsc{en}}}^{\ell}$. During inference, this vector is then scaled by a factor $\gamma$ and added to $\vh^{\ell}(x)$ to produce the modified activation, $\tilde{\vh}^{\ell}(x)$ as shown below:
\begin{equation}
\vv_{\text{\textsc{en}}}^{\ell} 
= \frac{1}{|\mathcal{S}|}
\sum_{x_{\text{\textsc{en}}}, x_{\text{\textsc{xx}}} \in \mathcal{S}}
\left( \vh^{\ell}(x_{\text{\textsc{en}}}) - \vh^{\ell}(x_{\text{\textsc{xx}}}) \right), \quad
\tilde{\vh}^{\ell}(x) = \vh^{\ell}(x) + \gamma\, \vv_{\text{\textsc{en}}}^{\ell}.
\label{eq:ensteer}
\end{equation}

\subsection{Experimental Settings}\label{sec:setup}

\paragraph{Evaluation Details} We measure the \tradeoff trade-off with multiple-choice datasets: \gmmlu and \blend, on six languages: Spanish (\textsc{es}), Indonesian (\textsc{id}), Korean (\textsc{ko}), Greek (\textsc{el}), Chinese (\textsc{zh}), and Arabic (\textsc{ar}).\footnote{For languages that are associated with multiple regions in \blend, we choose Spain for Spanish and South Korea for Korean.} As \blend does not include a development set, we create one by setting aside $200$ randomly chosen samples from the test data. The remaining data is what constitutes our true test set. We further split the development set into two: 100 samples for steering vector extraction and 100 for layer-wise analysis.\footnote{Detailed statistics of the two benchmarks are provided in the Appendix~\ref{sec:appendix_detail_benchmark}.} The model’s accuracy is determined by computing the log likelihood of each answer option and selecting the one with the highest probability. 

\paragraph{Model Training} We detail the training settings for each \cla approach below:
\begin{enumerate}
    \item \mist: We use $6$K English instruction–response pairs (first-turn only) from the OpenAssistant dataset \citep{openassistant23} and translate them into all six languages with Google Translate.\footnote{\url{https://translate.google.com}} We refer to this dataset as OpenAssistant\textsc{xx} and use it to train \mist{}.
    \item \midalign: We use OpenAssistant\textsc{xx} and \textsc{flores} \citep{flores200nature}, alternating between \textsc{sft} and \midalign{} loss, respectively. Layer $24$ is set as the middle layer for extracting representations to compute the alignment loss, following prior work.%
    \item \clo: We create preference pairs using the multi-way parallel OpenAssistant\textsc{xx} dataset and use $\lambda = 0.5$ and $\beta = 1.0$, following \cite{lee-etal-2025-cross}.
    \item \ensteer: While activation steering can be applied to any model, we default to the \base model unless noted. Activations are extracted using $100$ samples from \gmmlu's dev set. We set $\gamma$ to $2$. We use layer-wise Principal Component Analysis \citep[\textsc{pca}]{pca} analysis and identify layers $16$-$32$ to exhibit the highest overlap of hidden activations across languages which is necessary for steering to be effective. We apply \ensteer at layer $20$ based on the accuracy on the development set.
\end{enumerate}

\paragraph{Model Architecture} We conduct all our experiments using the Gemma3 $12$B pre-trained model \citep{gemma3_2025}, which consists of $48$ transformer layers. Gemma3 models have been trained on a vast amount of multilingual datasets, natively supporting $35$+ languages, which makes it a good candidate to evaluate cross-lingual transfer. All post-training models (\mist, \clo, \midalign) are trained on seven languages including English, for only one epoch, updating all parameters.\footnote{More details about post-training are in Appendix \ref{sec:appendix_detail_training}.}
 
\subsection{Mapping Cross-lingual Alignment to the Transfer-Localization Plane}\label{sec:mapping}

\begin{figure}[t!]
\centering
\begin{subfigure}[b]{0.50\textwidth}
    \includegraphics[width=\textwidth]{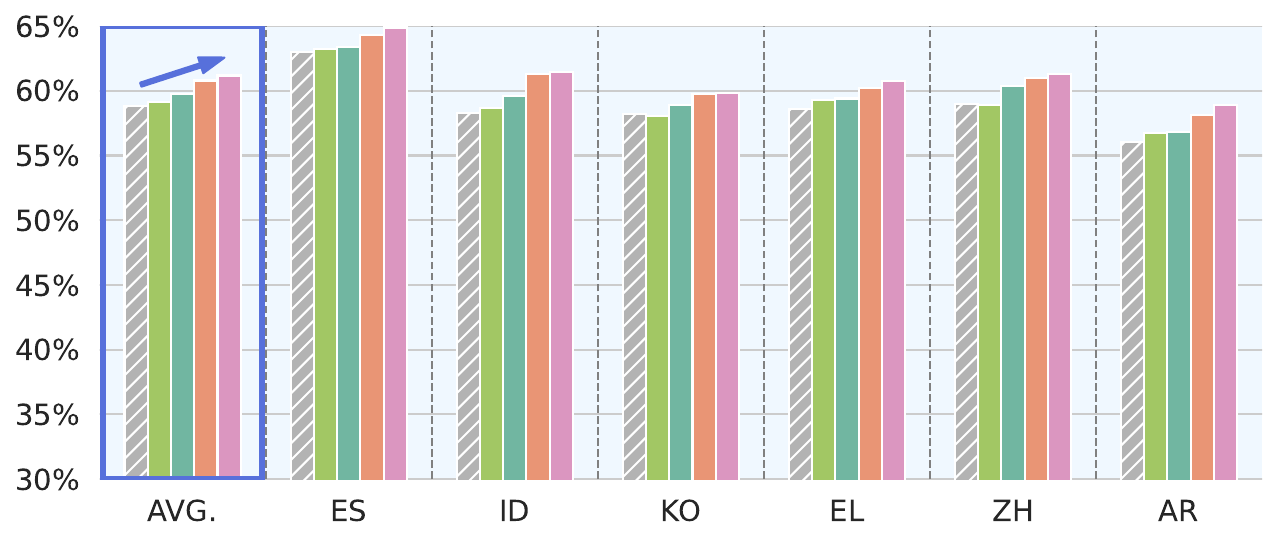}
\caption{\gmmlu accuracy.}
\label{fig:bar_globalmmlu}
\end{subfigure}
\begin{subfigure}[b]{0.49\textwidth}
    \includegraphics[width=\textwidth]{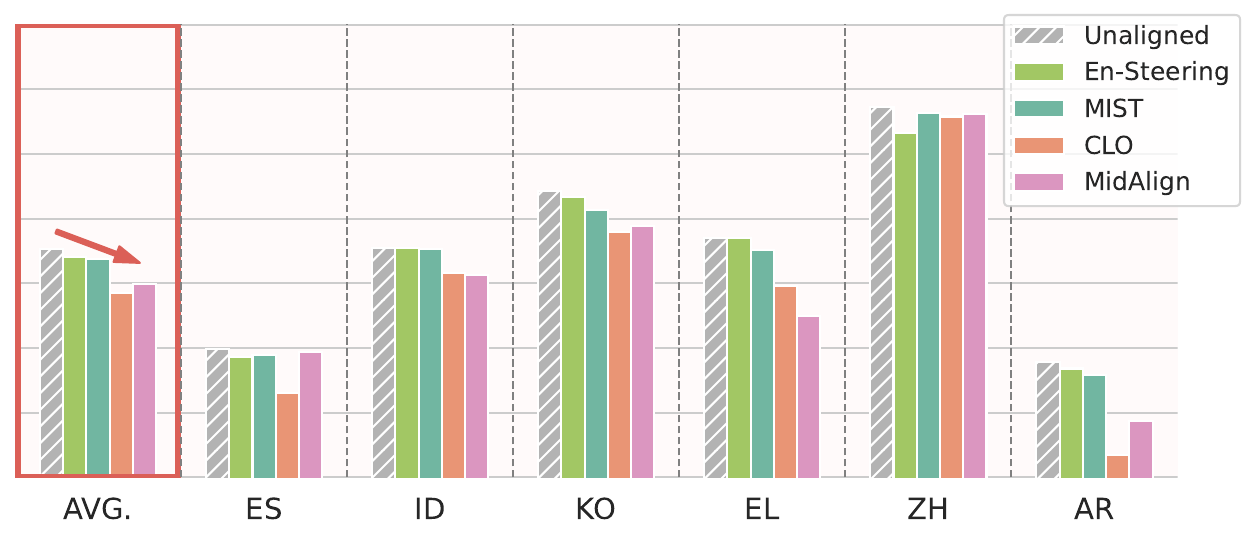}
\caption{\blend accuracy.}
\label{fig:bar_blend}
\end{subfigure}
\begin{subfigure}[b]{01\textwidth}
    \includegraphics[width=0.98\textwidth]{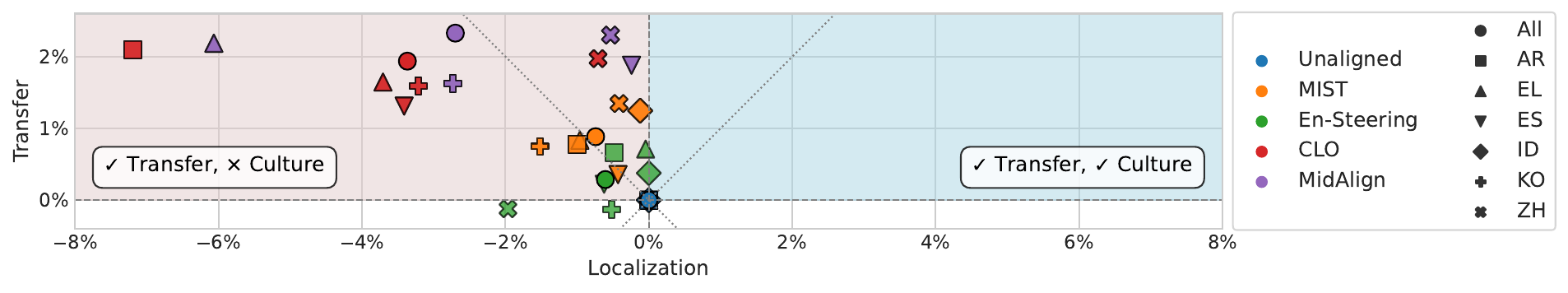}
\caption{Transfer-Localization trade-offs ($\Delta = \text{Acc}_{\text{\cla}} - \text{Acc}_{\text{\base}}$) on \gmmlu and \blend. }
\label{fig:pareto_nosteer}
\end{subfigure}
\caption{
Competing results of \cla approaches on knowledge transfer and cultural localization. \textbf{Improvements in \cla come at a \textit{consistent} cost of cultural localization across all languages.}
}
\label{fig:bar}
\end{figure}

Figure~\ref{fig:bar} presents performance of the pre-trained model (i.e., \base) and each of the \cla approaches on \gmmlu (universal knowledge) and \blend (culturally-adaptive knowledge) datasets.

\paragraph{\cla effectively improves knowledge transfer.} When evaluating universal knowledge transfer on \gmmlu (Fig. \ref{fig:bar_globalmmlu}), we observe that all \cla approaches generally improve performance over the \base baseline, across six non-English languages. However, the magnitude of this improvement varies. For instance, methods like \midalign(+$2.3$\%) and \clo (+$1.9$\%) consistently deliver the largest gains, suggesting that more explicit alignment is highly effective at bridging significant knowledge gaps. In contrast, \mist (+$0.9$\%) and \ensteer (+$0.3$\%) provide more modest, though still positive, gains. This finding is in complete alignment with results from prior work and indicates that when universal transfer is the target, all studied approaches are deemed successful.

\paragraph{\cla results in potential cultural erasure as suggested by the accuracy drop in \blend.} Results on the culturally adaptive \blend dataset (Fig. \ref{fig:bar_blend}) reveal the significant cost of this alignment. All alignment methods lead to a degradation in performance on culturally specific questions. This loss of nuance is particularly pronounced for the most effective transfer methods. For example, \clo, which showed strong gains on \gmmlu, consistently causes the most substantial performance drop (-$3.4$\%) on \blend across nearly all languages. This suggests that its aggressive representation alignment overwrites culturally specific information. Conversely, \mist, which enforces alignment more implicitly, induces the least amount of cultural erasure, preserving cultural knowledge more effectively than other methods. %

\paragraph{\cla exhibits \tradeoff tradeoffs.} To better show these competing outcomes, we plot the performance of each alignment method and language on a \tradeoff plane (Fig~\ref{fig:pareto_nosteer}). This plot positions each model-language pair based on its transfer gain (\gmmlu improvement, y-axis) against its cultural localization (\blend performance change, x-axis). The resulting frontier clearly illustrates the trade-off: methods that push further up (gaining transfer) invariably push further to the left (incurring erasure). A closer look at the frontier reveals distinct behaviors. The most aggressive alignment methods, \midalign (purple) and \clo (red), occupy the top-left region of the plot, while ``safer'' approaches with minimal cultural erasure constitute less powerful options for generalization, with certain languages such as Korean and Chinese even exhibiting degradation in transfer. Finally, this plot highlights that for many \cla methods, the cost of erasure outweighs the benefit of transfer. This establishes the central challenge for our next section: how to move beyond this frontier and achieve transfer without hurting cultural localization.

\subsection{The Internal Dynamics of Cross-lingual Alignment}\label{sec:cla_dynamics}

\begin{figure}[t!]
\centering
\begin{subfigure}[b]{0.49\textwidth}
    \includegraphics[width=\textwidth]{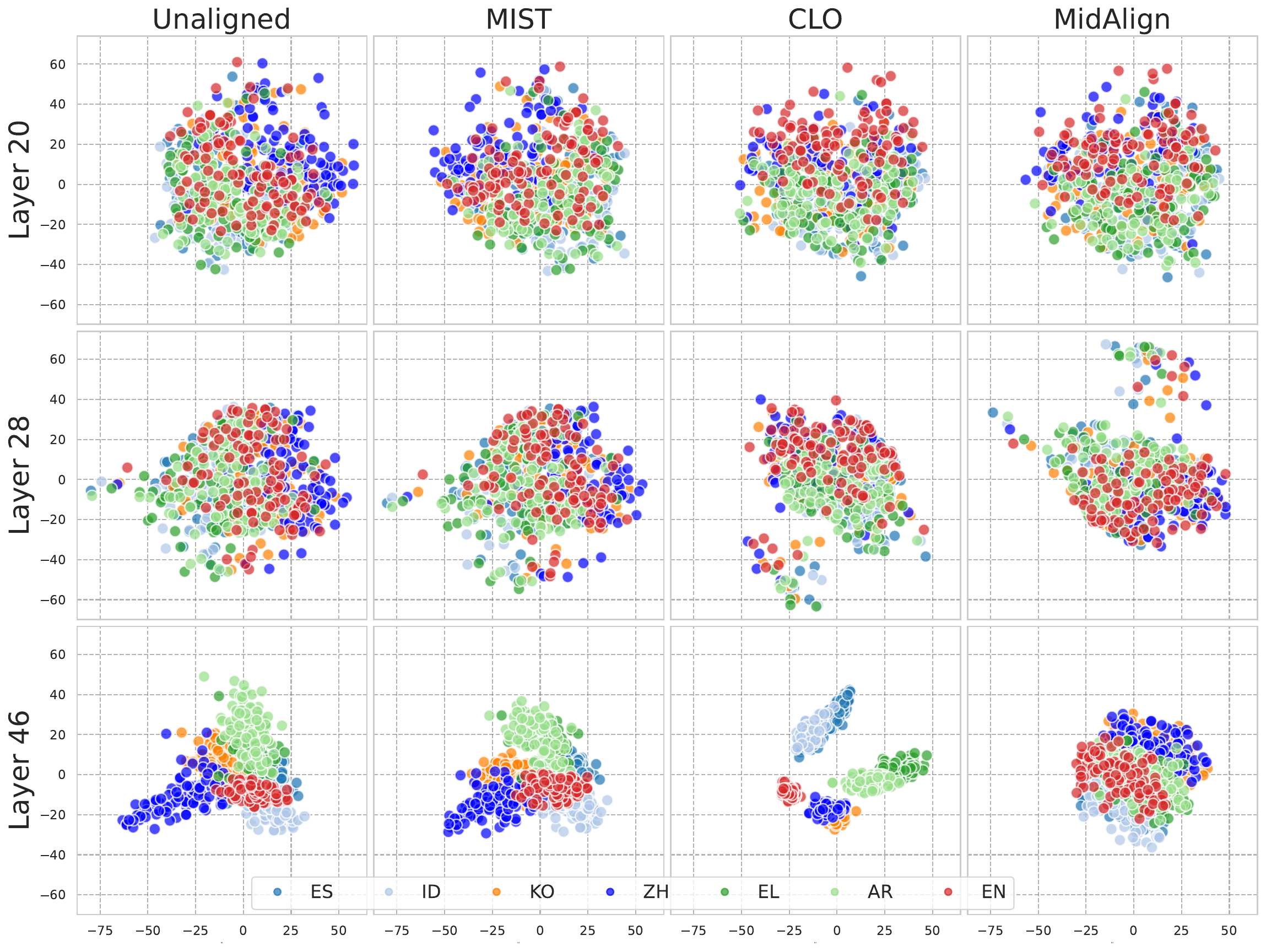}
\caption{\gmmlu}
\label{fig:gmmlu_pca}
\end{subfigure}
\begin{subfigure}[b]{0.49\textwidth}
    \includegraphics[width=\textwidth]{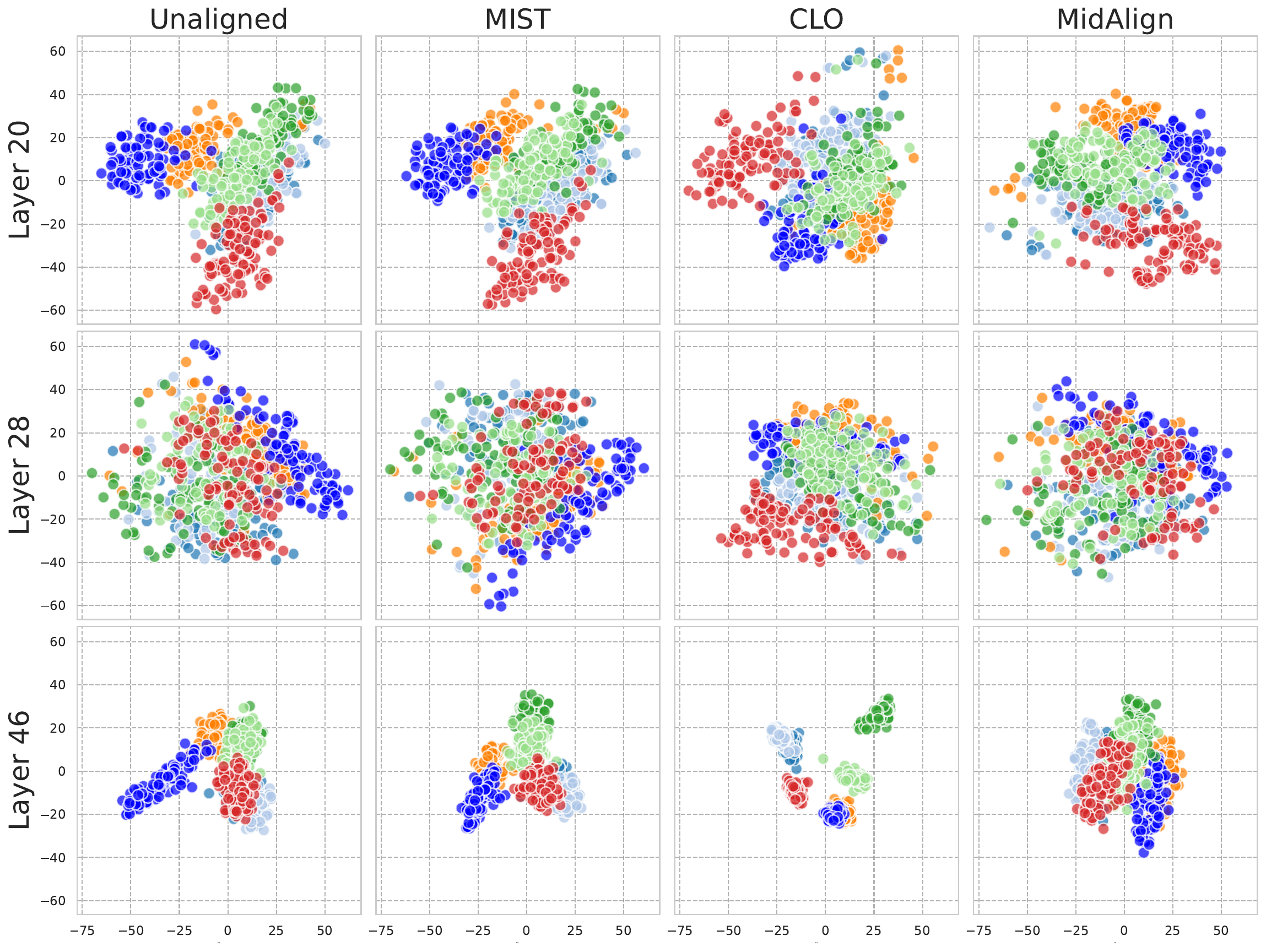}
\caption{\blend}
\label{fig:blend_pca}
\end{subfigure}
\caption{\pca projections of hidden representations across \base and \cla methods. As \cla methods are applied, languages cluster more tightly, signaling stronger convergence.
Yet, convergence differs by the nature of the datasets: \gmmlu merges starting in the middle layers, whereas \blend maintains separation until later stages, persisting even after \cla.}
\label{fig:smart_pca}
\end{figure}

How do \cla approaches alter the model's internal representation space? We analyzed \textsc{pca} projections of hidden states from various layers (middle: $20$, deep: $28$, outer: $47$). As shown in Figure~\ref{fig:smart_pca}, we compared the unaligned base model to three progressively aligned models---\mist, \clo, and \midalign---across the \gmmlu and \blend datasets.\footnote{A comprehensive set of plots for all layers can be found in the Appendix~\ref{sec:appendix_pca_mist}.}

Our analysis reveals two key dynamics. First, the alignment process differs significantly depending on the nature of the data. On the universal \gmmlu dataset, \cla methods successfully merge representations in the middle layers as intended. However, on the cultural \blend dataset, language representations remain largely separable in these same middle layers, with alignment only beginning to emerge deeper in the model. Surprisingly, this cultural separability persists even in the most stringently aligned models.
Second, regardless of the dataset, \cla methods (especially \clo and \midalign) induce a stronger representational convergence in the deeper layers (e.g., layer $28$).  
This raises the question: given the persistent representational differences on cultural data, could the associated performance losses on \blend be recovered using techniques like representation steering?

\section{Balancing Transfer and Cultural Erasure}\label{sec:mitigating}

If \cla suppresses a model's ability to use language as a cultural cue, is that knowledge permanently erased or merely inaccessible?
We start by exploring how existing activation steering techniques can be used to probe for localized knowledge (\S\ref{sec:loc_steer}). Then, 
we present a crucial finding from our representation analysis: knowledge transfer and cultural localization are optimally steered at different layers (\S\ref{sec:layerwise_steer}). Finally, we use this insight to better balance transfer and localization for \cla (\S\ref{sec:sur_steer}).

\subsection{Probing for Localized Knowledge with Activation Steering}\label{sec:loc_steer}

\begin{wraptable}{L}{0.6\textwidth} 
\centering
\caption{Transfer-Localization trade-offs for different steerings methods (applied on middle layer; avg. across langs).
}
\label{table:locsteer}
\begin{tabular}{lll} 
\toprule
\cla &
\gmmlu (\%) &
\blend (\%) \\
\midrule
\mist                          & $59.74$ &  $46.90$ \\
\quad + \ensteer               & $59.90$ \increase{0.16} & $46.45$ \decrease{0.45} \\
\quad + \locsteer              & $59.60$ \decrease{0.14} & $48.12$ \increase{1.22} \\

\bottomrule
\end{tabular}
\end{wraptable}

To investigate the extent to which localized knowledge remains accessible within aligned models, we adopt the localized activation steering method of \citet{locsteering25}.
Concretely, we use pairs of inputs with and without cultural context ($x_{\text{\textsc{con}}}$, $x_{\text{\textsc{decon}}}$) to derive a localizing vector~$\vv_{\text{\textsc{loc}}}^{\ell}$ (\locsteer), pushing the model toward local subspaces.\footnote{The localized steering vectors are extracted using the \blend development set described in Section~\ref{sec:setup}.} 
\begin{equation}
\vv_{\text{\textsc{loc}}}^{\ell} 
= \frac{1}{|\mathcal{S'}|}
\sum_{x_{\text{\textsc{con}}}, x_{\text{\textsc{decon}}} \in \mathcal{S'}}
\left( \vh^{\ell}(x_{\text{\textsc{con}}}) - \vh^{\ell}(x_{\text{\textsc{decon}}}) \right), \quad
\tilde{\vh}^{\ell}(x) = \vh^{\ell}(x) + \gamma\, \vv_{\text{\textsc{loc}}}^{\ell}.
\label{eq:locsteer}
\end{equation}

As shown in Table~\ref{table:locsteer},
applying \locsteer{} instead of \ensteer{} at the \mist's middle layer improves its ability to provide culturally situated responses. 
This indicates that cultural knowledge is not permanently erased but---at least to an extent--- suppressed, capable of being reactivated through targeted steering.
At the same time, this improvement on the cultural localization axis comes at a cost: universal transfer on \gmmlu degrades $0.3\%$ from \ensteer, \textbf{suggesting that transfer and cultural localization are not optimally co-located within the same model layers}.

\subsection{Knowledge Transfer and Cultural Localization Peak at Different Layers}\label{sec:layerwise_steer}

\begin{figure}[t!]
\centering
\begin{subfigure}[b]{0.235\textwidth}
    \includegraphics[width=\textwidth]{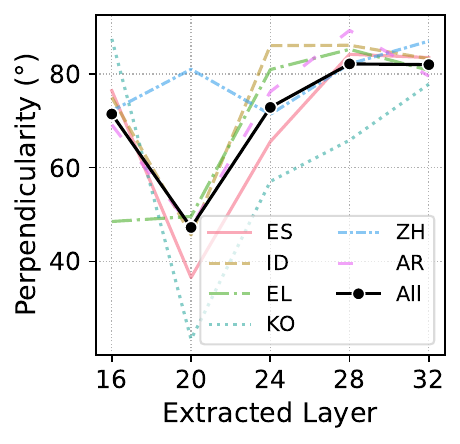}
\caption{Perpendicularity.}
\label{fig:layerwise_degree}
\end{subfigure}
\begin{subfigure}[b]{0.755\textwidth}
    \includegraphics[width=\textwidth]{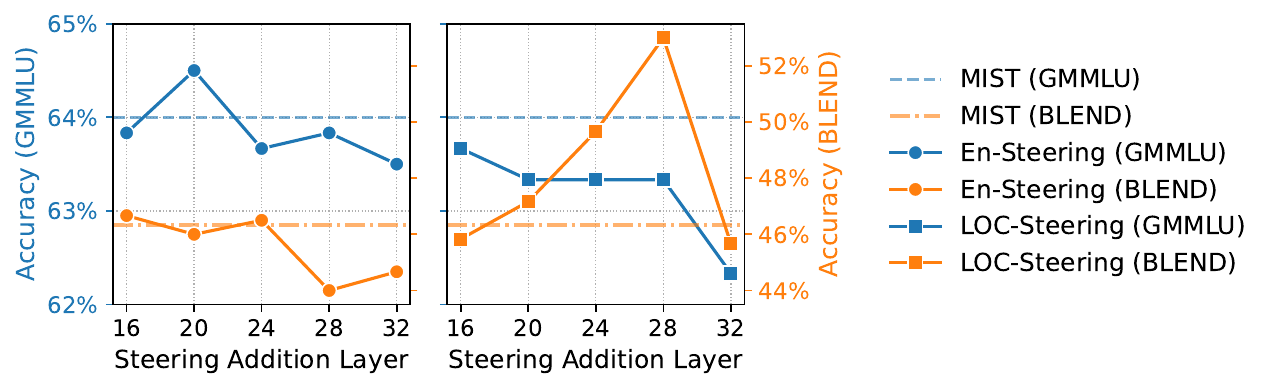}
\caption{Layer-wise performance on \gmmlu and \blend.}
\label{fig:layerwise_steering}
\end{subfigure}
\caption{Layer-wise analysis of \textsc{en-} and \locsteer on \mist for \gmmlu and \blend dev set (right) and perpendicularity between two kinds of vectors (left).
\textbf{Cultural localization is optimally located in deeper layers, where the \textsc{en}- and \textsc{loc} vectors are also most orthogonal to each other.}
}
\label{fig:layerwise}
\end{figure}

The above observations naturally prompt us to ask: where within a model's layers are cross-lingual transfer and cultural localization most effectively realized? 
We analyze the angle between the \ensteer{} and \locsteer{} vectors to identify layers where localization and transfer are disentangled. Intuitively, for these interventions to operate independently, their vectors should be orthogonal. As shown in Figure~\ref{fig:layerwise_degree}, this condition is met in the model's deeper layers, which approach orthogonality, but not in shallower layers where the vectors are closely aligned. Therefore, deeper layers (peaking at $28$) are optimal for applying localization steering with minimal interference, while shallower layers (e.g., $20$) risk conflicting signals.\footnote{More details of this analysis in Appendix~\ref{sec:appendix_perpen}.}

To systematically validate our observation, we analyze the layer-wise effects of applying \ensteer and \locsteer in isolation on a held-out dev set.\footnote{This is the second development set, which is distinct from the one used to extract steering vectors. Test set result is available in Appendix Figure~\ref{fig:layerwise_steering_test}, showing similar trend.} The results, shown in Figure~\ref{fig:layerwise_steering}, confirm our hypothesis from the angular analysis: while universal transfer (\ensteer) is most effective in the middle layers (peaking at layer $20$), cultural localization (\locsteer) performs optimally in the deeper layers (peaking at layer $28$). This finding echoes the hidden representation projection results in (\S\ref{sec:cla_dynamics}), where middle-layer representations across languages in \blend stay separated, making it unsuitable for effective steering. This layered separation has a critical practical implication: applying \locsteer at a deeper layer (e.g., $28$) significantly boosts performance on culturally situated questions without degrading universal transfer performance. Accordingly, we apply \locsteer at a deeper layer by default in subsequent experiments.

\subsection{Pushing the Transfer-Localization Frontier with Surgical Steering}\label{sec:sur_steer}

Motivated by our analysis above, we propose Surgical Steering (\sursteer): applying the \ensteer vector at an earlier layer~$\ell_{\text{\textsc{en}}}$ and the \locsteer vector at a deeper layer~$\ell_{\text{\textsc{loc}}}$ to have more controlled alignment.\footnote{Specifically, we simultaneously apply \ensteer on layer 20, and \locsteer on layer 28 for the models based on Gemma3 12B, which has $|L|$= 48 layers.}
Formally, our surgical intervention is defined as follows:
\begin{equation}
\tilde{\vh}_{l}(x)
= \vh_{l}(x)
+ \gamma \, \mathbf{1}_{l=\ell_{\text{\textsc{en}}}} \, \vv_{\text{\textsc{en}}}^{l}
+ \gamma \, \mathbf{1}_{l=\ell_{\text{\textsc{loc}}}} \, \vv_{\text{\textsc{loc}}}^{l},
\quad l \in \{1, \dots, |L|\}.
\end{equation}

\begin{figure}[t]
\centering
\begin{subfigure}[b]{0.6\textwidth}
    \includegraphics[width=\textwidth]{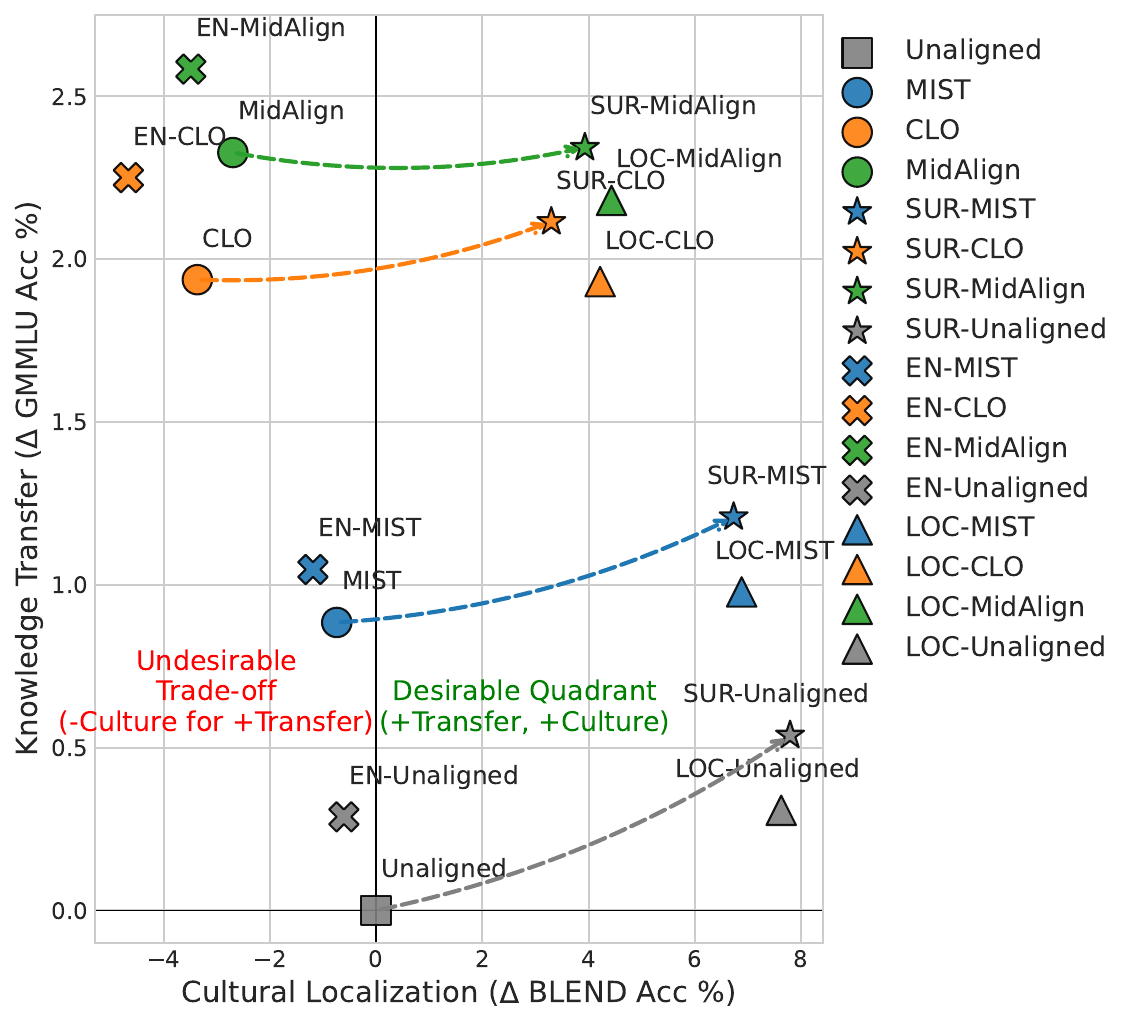}
\caption{Tranfer-Localization \cla trade-offs.}
\label{fig:pareto_steer}
\end{subfigure}
\begin{subfigure}[b]{0.39\textwidth}
    \includegraphics[width=\textwidth]{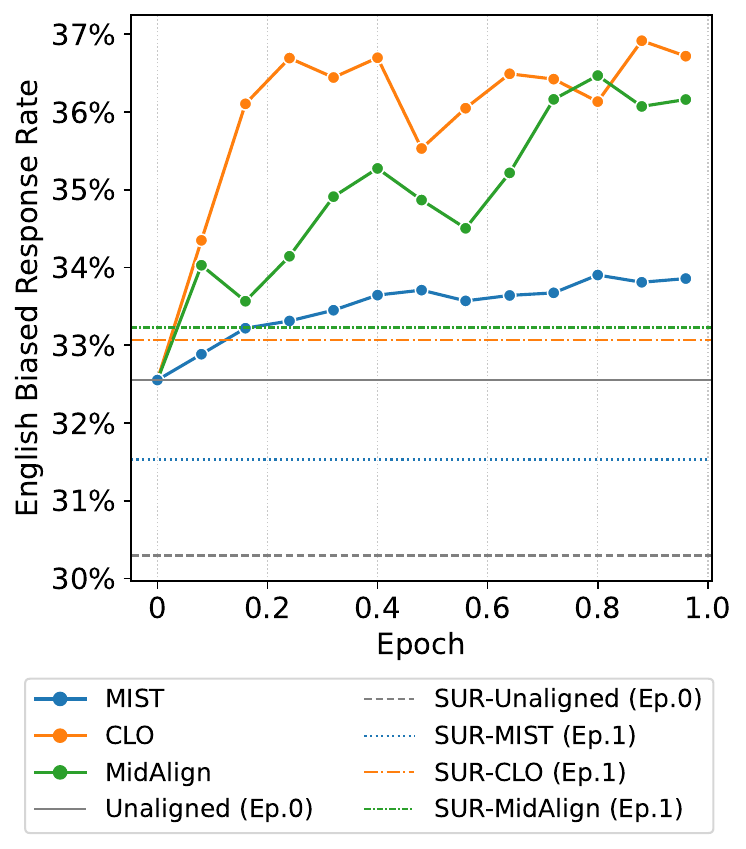}
\caption{\cla English-centric bias.}
\label{fig:ckpt_trend}
\end{subfigure}
\caption{Left: Trade-offs between transfer and localization with steering methods. Both \ensteer and \locsteer are applied to \mist. \sursteer{} is applied on top of different post-training methods (circles), indicated by the same color or by the connecting gray dotted line.
Right: Tracking English-bias of post-training \cla methods and the impact of \sursteer on all approaches.}
\label{fig:pareto_and_ckpt_trend}
\end{figure}

\paragraph{Superiority of \sursteer{}.} Figure~\ref{fig:pareto_steer} shows that \sursteer{} achieves a more favorable trade-off than applying either \ensteer{} or \locsteer{} alone. It surpasses \ensteer{} in cross-lingual transfer while simultaneously improving cultural localization. This demonstrates that combining steering at distinct layers provides finer control over the alignment process, pushing the Pareto frontier towards a more optimal state.
Language-wise results are in Appendix Table~\ref{tab:allvalues}.

\paragraph{General Steerability of \cla Models.} Our experiments also reveal a broader insight: all tested \cla approaches remain steerable. As shown by the orange and green stars in Figure~\ref{fig:pareto_steer}, applying \sursteer{} to models already trained with \midalign{} and \clo{} yields further improvements in both transfer and localization. The gain for \midalign{} is smaller, however, suggesting a saturation effect in models that already possess high transfer capabilities.

\label{sec:appendix_posttrain_trend}
\paragraph{Addressing English Bias with \sursteer{}.}
Given the hypothesis that multilingual \llms improve cross-lingual transfer by aligning representations towards English~\citep{wendler-etal-2024-llamas}, we assess whether surgical steering improves cultural localization by suppressing the English-centric responses. We extract all the queries from \blend ($\sim$40\%) that include an answer associated with English-speaking countries (\textsc{us}/\textsc{uk}) and measure the proportion of times the model selects this option as an answer for a non-English query. Figure~\ref{fig:ckpt_trend} shows that first, as the training progress, the model's tendency to select the English option increases across all \cla methods (\mist{}, \clo{}, \midalign{}), validating this hypothesis.\footnote{We also show the accuracy trends for all \cla approaches in the Appendix Figure~\ref{fig:ckpt_trend_accuracy}.} However, applying \sursteer{} to all approaches reduces this bias significantly (up to 4\%), showing its effectiveness in steering the model away from English-centric responses. We note that the bias, however, is not completely removed as even when steering the \base model, the accuracy of English responses remains high at $30.2$\%.
\

\paragraph{The Irrecoverable Trade-off.} Critically, the fundamental trade-off persists. Models with stronger alignment, such as \midalign{} and \clo{}, are less responsive to steering than the \base or \mist{} models (Fig.~\ref{fig:pareto_steer}). This indicates that while cultural knowledge is partially recoverable, some cultural nuances are irrevocably lost during the alignment process. The same applies to  English-biased responses (Fig.~\ref{fig:ckpt_trend}): although steering alleviates the bias to some extent, the post-trained \cla models never fully return to the point of the \base baseline, indicating inherent limits to what the steering can recover.

\section{Conclusion}
In this work, we address the critical trade-off between knowledge transfer and cultural localization in cross-lingual alignment. We introduce a holistic framework to systematically measure this trade-off, quantifying not only the gains in universal knowledge transfer but also the loss of cultural specificity. Our empirical analysis confirms that existing alignment methods consistently improve transfer at the direct cost of cultural localization.
To mitigate this, we propose a simple yet effective method using controllable activation steering. We demonstrate that by disentangling universal and localized steering at different, optimal layers, we can improve performance on culturally situated tasks without compromising transfer. This reveals a key insight: cultural knowledge is not permanently erased by alignment but is instead suppressed, making it partially recoverable through targeted, layer-specific interventions. Nevertheless, our findings suggest that some cultural nuances are irretrievably lost, highlighting a persistent and perhaps unavoidable cost of the cross-lingual alignment process.

\bibliography{iclr2026_conference}
\bibliographystyle{iclr2026_conference}

\section*{The Use of Large Language Models}
We use \llm to partially refine or polish writing at the sentence level (e.g., fixing grammar, re-wording sentences)

\appendix
\newpage
\section{Additional Experimental Details}\label{sec:appendix_detail}

\subsection{Training Details}\label{sec:appendix_detail_training}
We use 16 TPUv4 for post-training and set the batch size to 16 for all experiments.
We use a maximum sequence length of 640 tokens.
The peak learning rate is set to 5e-5 for \mist and 5e-4 for the rest.
All our implementations are based on \texttt{Flax} \citep{flax2020github}, a neural network library for \texttt{jax} \citep{jax2018github}.

\subsection{Benchmarks and Evaluation}\label{sec:appendix_detail_benchmark}

For creating decontextualized \blend queries, we prompt Gemini 2.5 Flash with the instruction provided in Figure~\ref{fig:prompt_decontext}. Outputs are automatically checked against the originals, and cases with excessive reduction or unintended content changes are filtered and re-processed with Gemini 2.5 Pro.
Since \blend has English multiple choice options, we translate the non-English choices provided in the dataset with Gemini 2.5 Flash using the prompt shown in Figure~\ref{fig:prompt_option}.
For cases where the translation output does not match the predefined format, we re-translate using Gemini 2.5 Pro.

The prompt used for the evaluation of multiple choice questions is shown in Figure~\ref{fig:prompt_mcq}.

For the development set, we randomly select $200$ samples: from the development split of \gmmlu and from the original set of \blend. To extract the \ensteer and \locsteer vectors, we use $100$ samples (Dev1), reserving the remaining $100$ for layer-wise analysis to determine the optimal layer (Dev2). Detailed statistics for both benchmarks are shown in Table~\ref{tab:dataset}.
\begin{table}[h]
\centering
\caption{\gmmlu and \blend Statistics}
\label{tab:dataset}
\small
\begin{tabular}{clccclcccc}
\toprule
& & \multicolumn{3}{c}{\gmmlu} & \multicolumn{5}{c}{\blend} \\
\cmidrule(r){3-5} \cmidrule(l){6-10}
Code&  Language  &  Dev1  & Dev2 &  Test   &  Region   &  Extracted  &  Dev1 & Dev2 & Test \\
\midrule
\textsc{es} & Spanish & 100 & 100 & 14042 & Spain & 19280 & 100 & 100 & 19080 \\
\textsc{id} & Indonesian & 100 & 100 & 14042 & Indonesia & 18417 & 100 & 100 & 18217 \\
\textsc{ko} & Korean & 100 & 100 & 14042 & South Korea & 21439 & 100 & 100 & 21239 \\
\textsc{el} & Greek & 100 & 100 & 14042 & Greece & 20383 & 100 & 100 & 20183 \\
\textsc{zh} & Chinese & 100 & 100 & 14042 & China & 20410 & 100 & 100 & 20210 \\
\textsc{ar} & Arabic & 100 & 100 & 14042 & Algeria & 20364 & 100 & 100 & 20164 \\
\bottomrule
\end{tabular}
\end{table}

\begin{figure}[h!] 
\centering
\begin{lstlisting}[style=mypromptstyle]]
{question}
Without any explanation, choose only one from the given alphabet choices (e.g., A, B, C, D).
A. {option_a}
B. {option_b}
C. {option_c}
D. {option_d}
Answer:
\end{lstlisting}
\caption{The prompt template used for our multiple-choice question experiments.}
\label{fig:prompt_mcq}
\end{figure}

\begin{figure}[h!]  
\centering
\begin{lstlisting}[style=mypromptstyle]]
Analyze the following list of question. Identify the shared, core question by removing the specific location (e.g., "in US", "in UK", "in West Java") from the end of each sentence.
- Remove the specific context to create a natural, generalized question.
- Do not put [country] in it just remove the country name and make it natural.
- Do not paraphrase the original input. Just try to minimally remove the context from the input.
Provide only this single, decontextualized question as the output.
Input: {Question}
Output:
\end{lstlisting}
\caption{The prompt template used for decontextualizing a query.}
\label{fig:prompt_decontext}
\end{figure}

\begin{figure}[h!]  
\centering
\begin{lstlisting}[style=mypromptstyle]]
You are a professional translator, translating from English to {language} spoken in {country_name}.
Translate the given list of English keywords e.g. [key1, key2,...] and output in a dictionary format e.g. {{key1: translation1, key2: translation2, ...}}.
- A key that represents numerical data, a date, or a time (e.g., "123", "1,000", "10:30", "12/25") MUST be copied to its value instead of being translated.
- All other keys should be translated.
- All translation values MUST be a single string.
- If a hint is provided below for a specific keyword, you MUST use one of the suggested translations.
{hint_phrase}
List of keywords: {options}
Do NOT include any explanatory text, comments, or markdown formatting.
\end{lstlisting}
\caption{The prompt template used for translating options.}
\label{fig:prompt_option}
\end{figure}

\section{Additional Analysis}
\begin{figure}[ht!]
\centering
    \includegraphics[width=0.8\textwidth]{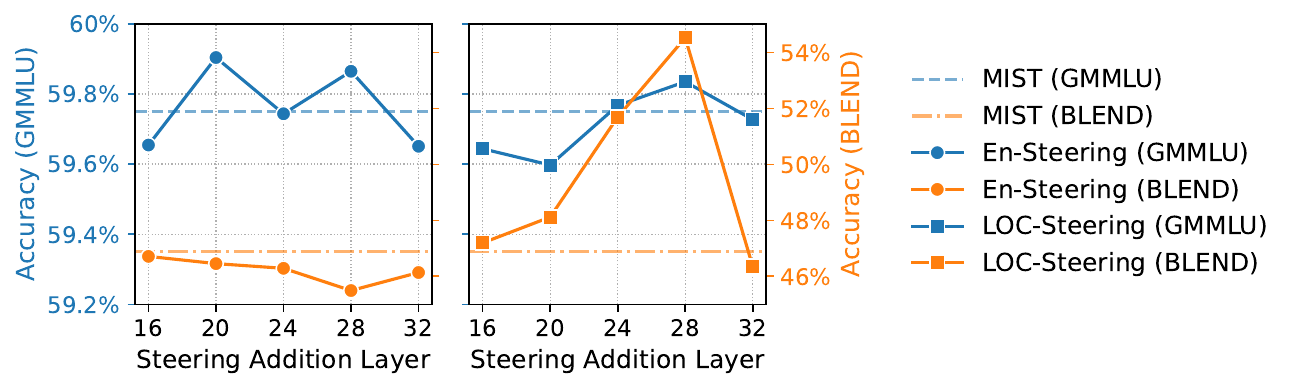}
\caption{
Layer-wise performance of \ensteer and \locsteer of \mist on \gmmlu and \blend test set. We observe that cultural steering is optimally located at deeper layers. The trends are similar in the development set (Figure~\ref{fig:layerwise_steering}).
}
\label{fig:layerwise_steering_test}
\end{figure}

\subsection{Perpendicularity Analysis Between \ensteer and \locsteer}\label{sec:appendix_perpen}
\begin{figure}[h]
\begin{center}
    \includegraphics[width=0.5\linewidth]{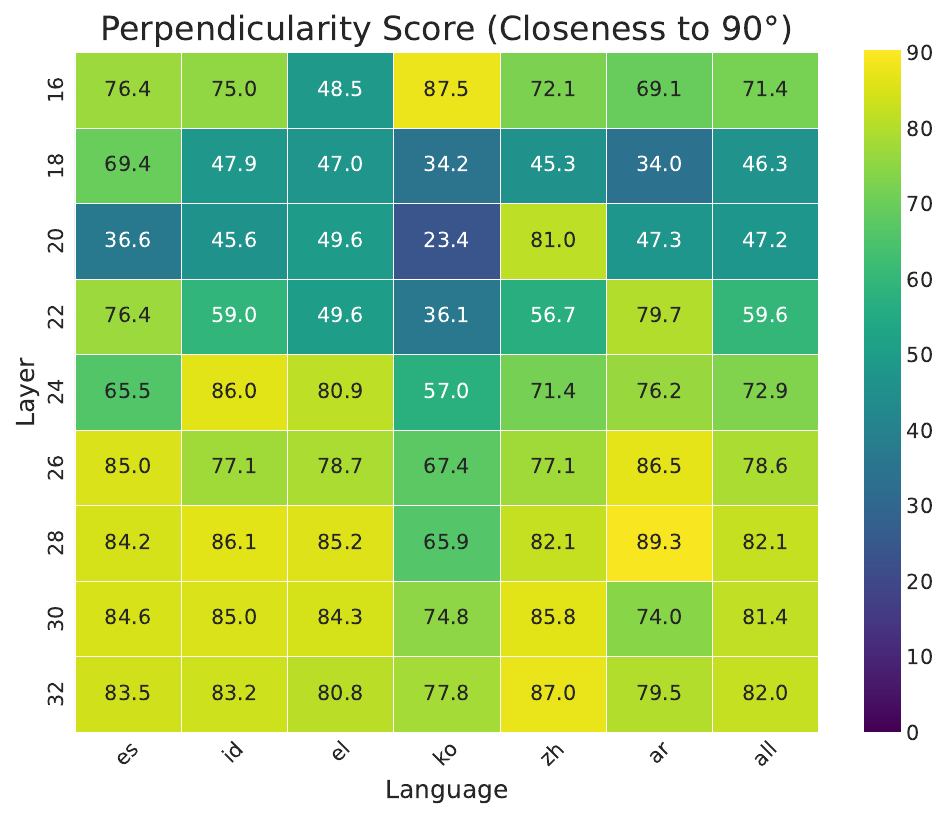} 
\end{center}
\caption{
Layer-wise perpendicularity analysis between \ensteer and \locsteer
}
\label{fig:perpendicularity_langs}
\end{figure}

We quantify perpendicularity as the deviation of the inter-vector angle from orthogonality between \ensteer vector and \locsteer extracted from each layer.
More specifically, we calculate a perpendicularity score $S^{\ell}_{\text{\textsc{per}}}$ between $\vv_{\text{\textsc{en}}}^{\ell}$ and $\vv_{\text{\textsc{loc}}}^{\ell}$ at layer $\ell$ based on the closeness of the vector angle to 90 degrees as follows:
\begin{equation} %
 S^{\ell}_{\text{\textsc{per}}} = 90 - \left| \left( \frac{180}{\pi} \arccos\left(\frac{\vv_{\text{\textsc{loc}}}^{\ell} \cdot \vv_{\text{\textsc{en}}}^{\ell}}{\|\vv_{\text{\textsc{loc}}}^{\ell}\| \|\vv_{\text{\textsc{en}}}^{\ell}\|}\right) \right) - 90 \right|.
\end{equation}
A score of 90 means the vectors are perfectly perpendicular (90°), and a score of 0 means the vectors are perfectly parallel (0° or 180°).
Overall, shallower layers exhibit lower perpendicularity between \ensteer{} and \locsteer{}, whereas deeper layers approach closer to orthogonality (Figure~\ref{fig:perpendicularity_langs}).
This implies that additional localization interventions can operate with minimal interference in deeper layers (peaking at 28), in contrast to shallower layers  where the effects of transfer and localization are more entangled (as seen in layer 20).

\subsection{Layer-wise PCA Analysis Across Cross-Lingual Alignment}\label{sec:appendix_pca_mist}
\begin{figure}[h]
\begin{center}
    \includegraphics[width=1\linewidth]{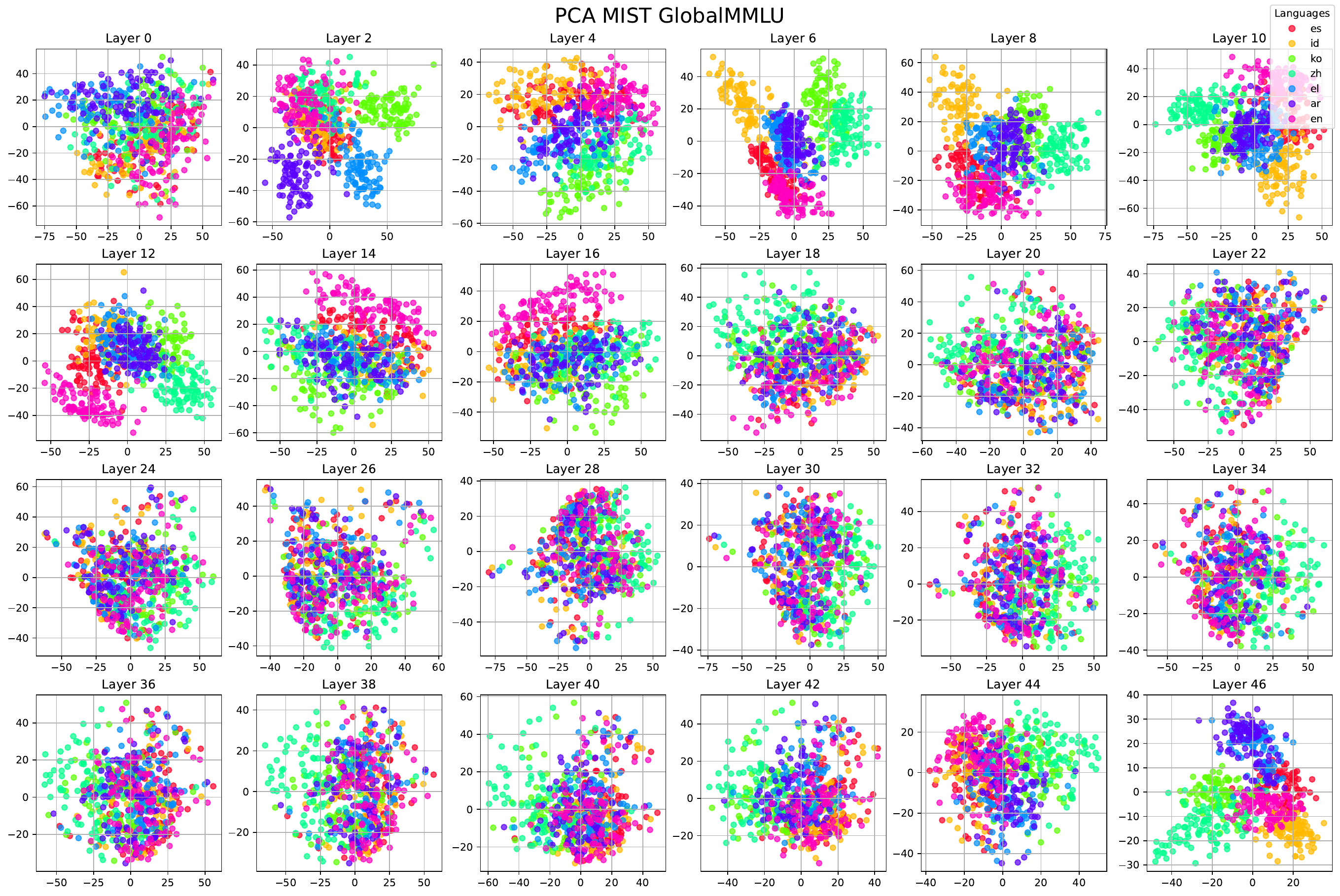} 
\end{center}
\caption{
Layer-wise \textsc{pca} plots of \mist with activations from \gmmlu samples. Each color represents a different language.
}
\label{fig:pca_mist}
\end{figure}

\begin{figure}[h]
\begin{center}
    \includegraphics[width=1\linewidth]{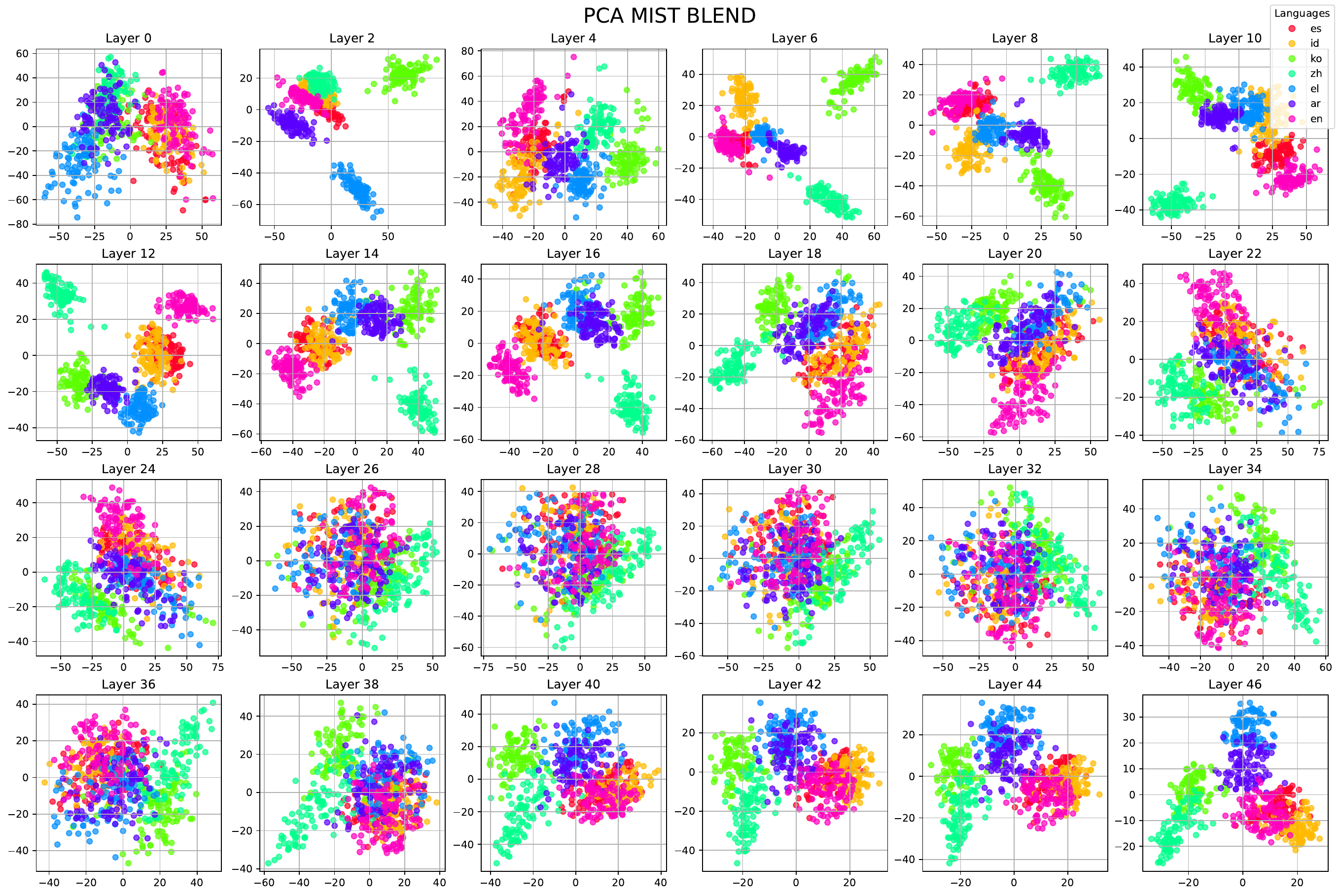} 
\end{center}
\caption{
Layer-wise \textsc{pca} plots of \mist with activations from \blend samples. Each color represents a different language.
}
\label{fig:pca_mist_blend}
\end{figure}

Following \citet{lim2025langlatenthinders}, we conduct a layer-wise \pca analysis with the extracted activations. In Figure~\ref{fig:pca_mist}, each color represents a different language activation extracted from the \gmmlu samples. Unlike the early or late layers, where languages appear more easily separated, the middle layers (16–32) show better overlaps across languages. We focus our analysis on these layers, where the representations become more tightly clustered, suggesting that steering vectors are most effective in this region. %
A \pca plot for \blend is in Figure~\ref{fig:pca_mist_blend}.


\subsection{Layer-wise PCA Analysis on Activation Steering}\label{sec:appendix_pca_steer}
\begin{figure}[h]
\begin{center}
    \includegraphics[width=1\linewidth]{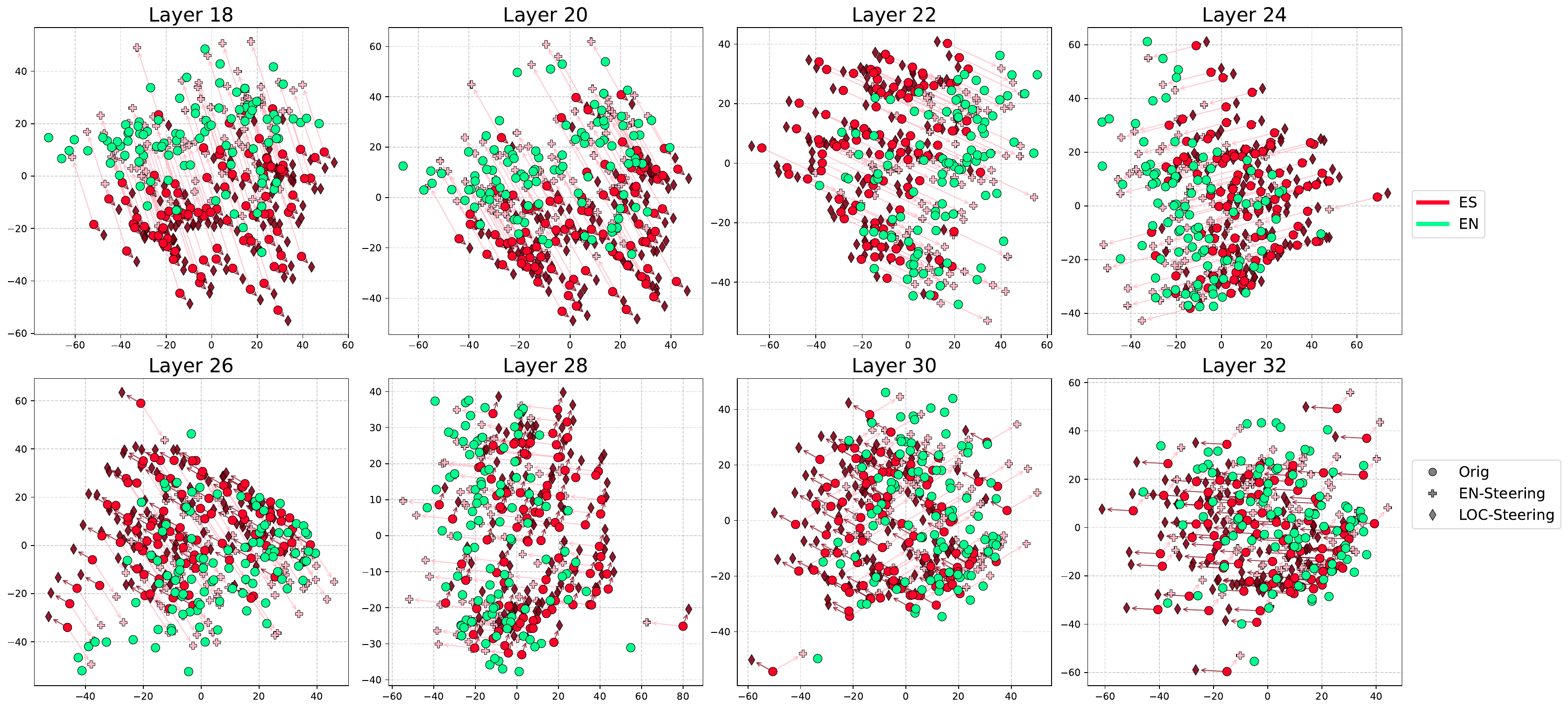} 
\end{center}
\caption{
Layer-wise \textsc{pca} plots of \mist. Each color represents a different language. 
\pca plots of Spanish and English hidden representations from \mist and its steering vectors.
By definition, \ensteer shifts Spanish vectors toward the English cluster, while \locsteer directs them toward localized subspaces.
}
\label{fig:pca_steer}
\end{figure}

Next, we examine how activation steering alters the geometry of hidden representations across layers.
Figure~\ref{fig:pca_steer} shows \pca projections of Spanish and English representations from the \mist model (Orig, circle), along with their transformations under \ensteer (cross) and \locsteer (diamond) interventions.
By definition, \ensteer shifts representations toward English, consistently pushing Spanish vectors closer to the English cluster.
In the earlier layers, the directions of the original English and localized representations often exhibit parallel or counter-parallel tendencies.
However, starting around layer 26 and becoming especially evident at layers 28 and 30, the vectors reveal a near-perpendicular relationship.
Even in this 2D projection, this perpendicularity is clearly observable, consistent with our earlier findings in Appendix~\ref{sec:appendix_perpen}.

\begin{figure}[t!]
\centering
\begin{subfigure}[b]{0.50\textwidth}
    \includegraphics[width=\textwidth]{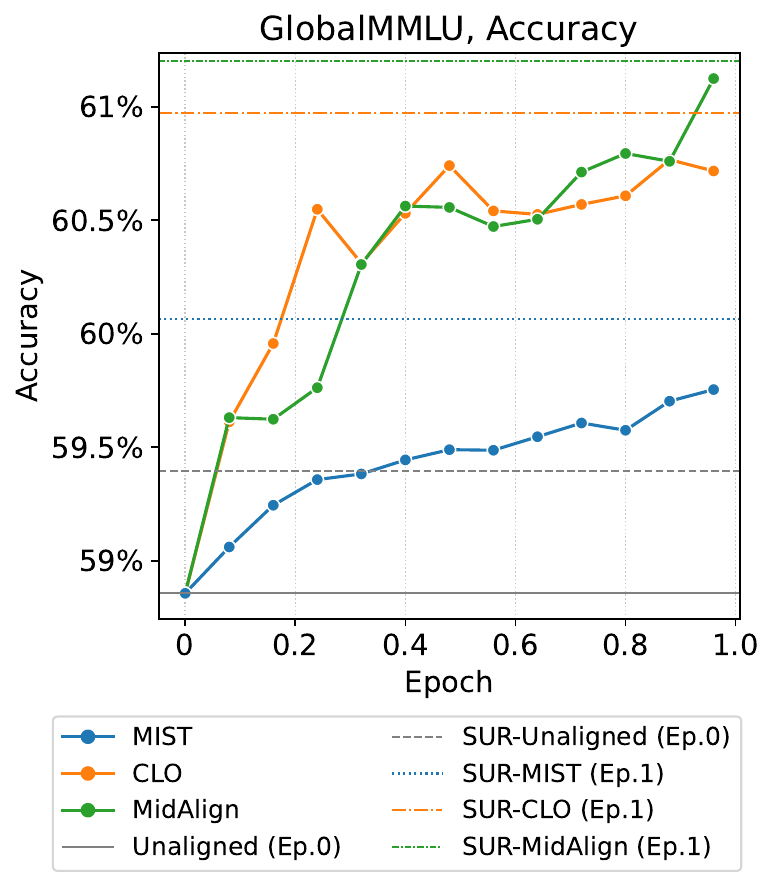}
\caption{\gmmlu}
\label{fig:ckpt_trend_gmmlu}
\end{subfigure}
\begin{subfigure}[b]{0.48\textwidth}
    \includegraphics[width=\textwidth]{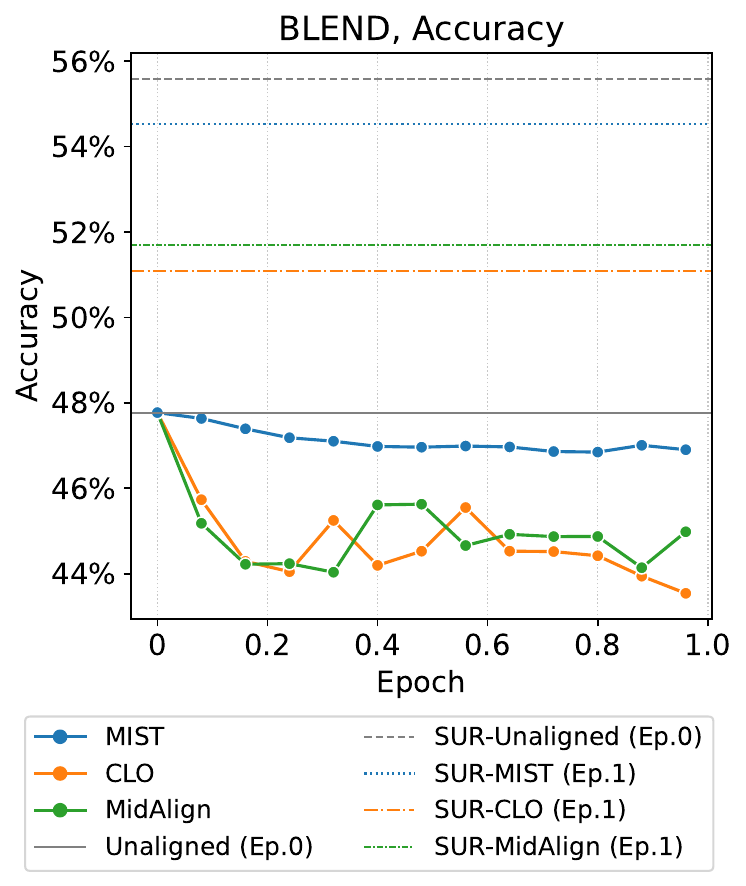}
\caption{\blend}
\label{fig:ckpt_trend_blend}
\end{subfigure}
\caption{Learning dynamics of post-training \cla methods across alignment epochs on \gmmlu and \blend.}
\label{fig:ckpt_trend_accuracy}
\end{figure}

\begin{table}[h]
\centering
\small
\caption{Accuracy (\%) of all \cla methods on \gmmlu and \blend test set. $^*$ denotes steering applied on a middle layer (20) and $^{**}$ on a deeper layer (28). \sursteer simultaneously apply \ensteer on the middle layer and \locsteer on the deeper layer.}
\label{tab:allvalues}
\begin{tabular}{lccccccc}
\toprule
\gmmlu (\%) & All     & \textsc{es}      & \textsc{id}      & \textsc{ko}      & \textsc{el}      & \textsc{zh}      & \textsc{ar}      \\
\midrule
\base                             & 58.86 & 63.01 & 58.32 & 58.19 & 58.57 & 59.00 & 56.05 \\
\quad + \ensteer$^*$                     & 59.14 & 63.23 & 58.70 & 58.06 & 59.29 & 58.88 & 56.71 \\
\quad + \locsteer$^*$     & 58.87 & 62.87 & 58.28 & 58.18 & 58.47 & 59.32 & 56.12 \\
\quad + \locsteer$^{**}$     & 59.16 & 63.28 & 58.94 & 58.14 & 58.39 & 59.69 & 56.55 \\
\quad + \sursteer                    & 59.39 & 63.45 & 59.15 & 58.50 & 59.22 & 59.27 & 56.79 \\
\mist                             & 59.74 & 63.37 & 59.56 & 58.94 & 59.41 & 60.35 & 56.82 \\
\quad + \ensteer$^*$                     & 59.90 & 63.75 & 59.69 & 58.79 & 59.87 & 60.03 & 57.29 \\
\quad + \locsteer$^*$     & 59.60 & 63.30 & 59.28 & 58.83 & 59.41 & 60.16 & 56.60 \\
\quad + \locsteer$^{**}$     & 59.83 & 63.65 & 59.88 & 58.48 & 59.18 & 60.55 & 57.26 \\
\quad + \sursteer                    & 60.07 & 64.01 & 60.15 & 58.99 & 59.79 & 60.09 & 57.36 \\
\clo                              & 60.79 & 64.32 & 61.33 & 59.78 & 60.22 & 60.97 & 58.14 \\
\quad + \ensteer$^*$                      & 61.11 & 64.93 & 61.63 & 60.08 & 60.76 & 60.77 & 58.46 \\
\quad + \locsteer$^*$      & 60.79 & 64.48 & 61.29 & 59.72 & 60.40 & 60.93 & 57.91 \\
\quad + \locsteer$^{**}$      & 60.79 & 64.00 & 61.16 & 59.93 & 60.52 & 60.71 & 58.40 \\
\quad + \sursteer                     & 60.97 & 64.44 & 61.28 & 59.91 & 60.74 & 60.68 & 58.79 \\
\midalign                         & 61.18 & 64.89 & 61.44 & 59.81 & 60.76 & 61.30 & 58.90 \\
\quad + \ensteer$^*$                 & 61.44 & 65.07 & 61.69 & 60.22 & 61.22 & 61.20 & 59.24 \\
\quad + \locsteer$^*$ & 61.22 & 64.96 & 61.45 & 60.04 & 60.78 & 61.23 & 58.86 \\
\quad + \locsteer$^{**}$ & 61.04 & 64.44 & 61.38 & 59.46 & 60.33 & 61.64 & 58.97 \\
\quad + \sursteer                & 61.20 & 64.83 & 61.39 & 59.55 & 60.48 & 61.62 & 59.33 \\
\midrule
\blend (\%) & All     & \textsc{es}      & \textsc{id}      & \textsc{ko}      & \textsc{el}      & \textsc{zh}      & \textsc{ar}      \\
\midrule
\base                             & 47.64 & 39.91 & 47.75 & 52.17 & 48.53 & 58.60 & 38.91 \\
\quad + \ensteer$^*$                     & 47.04 & 39.29 & 47.75 & 51.65 & 48.48 & 56.64 & 38.42 \\
\quad + \locsteer$^*$     & 48.71 & 40.39 & 48.76 & 54.51 & 49.43 & 60.02 & 39.13 \\
\quad + \locsteer$^{**}$     & 55.27 & 48.97 & 54.25 & 64.90 & 57.10 & 65.14 & 41.27 \\
\quad + \sursteer                    & 55.44 & 48.60 & 55.30 & 65.07 & 57.77 & 65.24 & 40.65 \\
\mist                             & 46.90 & 39.48 & 47.63 & 50.65 & 47.56 & 58.18 & 37.90 \\
\quad + \ensteer$^*$                     & 46.45 & 39.33 & 47.92 & 50.24 & 47.28 & 56.62 & 37.35 \\
\quad + \locsteer$^*$     & 48.12 & 39.82 & 48.44 & 53.32 & 48.60 & 60.07 & 38.46 \\
\quad + \locsteer$^{**}$     & 54.53 & 48.19 & 53.89 & 63.59 & 56.67 & 64.46 & 40.36 \\
\quad + \sursteer                    & 54.37 & 47.56 & 54.48 & 63.20 & 56.78 & 64.60 & 39.63 \\
\clo                              & 44.28 & 36.50 & 45.78 & 48.96 & 44.82 & 57.89 & 31.71 \\
\quad + \ensteer$^*$                      & 42.98 & 35.69 & 44.97 & 47.22 & 43.11 & 55.45 & 31.44 \\
\quad + \locsteer$^*$      & 45.41 & 37.07 & 47.13 & 51.22 & 45.88 & 58.90 & 32.27 \\
\quad + \locsteer$^{**}$      & 51.86 & 44.95 & 51.75 & 59.81 & 55.40 & 62.37 & 36.89 \\
\quad + \sursteer                     & 50.94 & 44.19 & 50.68 & 58.51 & 54.48 & 61.51 & 36.29 \\
\midalign                         & 44.95 & 39.67 & 45.65 & 49.44 & 42.46 & 58.07 & 34.39 \\
\quad + \ensteer$^*$                 & 44.15 & 39.54 & 44.66 & 48.87 & 42.08 & 56.26 & 33.52 \\
\quad + \locsteer$^*$ & 46.52 & 40.38 & 47.94 & 51.57 & 44.16 & 59.59 & 35.47 \\
\quad + \locsteer$^{**}$ & 52.07 & 47.32 & 51.45 & 59.40 & 53.13 & 62.75 & 38.40 \\
\quad + \sursteer                & 51.57 & 46.60 & 51.15 & 58.47 & 53.39 & 62.39 & 37.43 \\
\bottomrule
\end{tabular}
\end{table}

\end{document}